%% file: mainV4.tex
\documentclass[conference,a4paper]{IEEEtran}
\IEEEoverridecommandlockouts

\usepackage[hidelinks]{hyperref}
\usepackage[cmex10]{amsmath}%American Math Society(AMS) math formatting
\usepackage{amssymb,amsfonts}%AMS extra symbols and fonts
\usepackage{dblfloatfix}%fix double column figure ordering and placement

\usepackage[ruled,vlined]{algorithm2e}
\usepackage{graphicx}
\graphicspath{{Figures/PDF/}{Figures/PNG/}}

\usepackage{booktabs}
\usepackage{siunitx}
\usepackage[numbers,compress]{natbib}
\usepackage{texnames}
\usepackage{bm,bbm}
\usepackage{orcidlink}

\begin{document}

\title{Physics-Conditioned Synthesis of Internal Ice-Layer Thickness for Incomplete Layer Traces
\thanks{This work is supported by NSF BIGDATA awards (IIS-1838230, IIS-2308649), NSF Leadership Class Computing awards (OAC-2139536), NSF PFI award (2423211), IBM, and Amazon. This work used Jetstream2 at Indiana University through allocation CIS250588 from the Advanced Cyberinfrastructure Coordination Ecosystem: Services \& Support (ACCESS) program, which is supported by National Science Foundation grants \#2138259, \#2138286, \#2138307, \#2137603, and \#2138296}
\thanks{Copyright 2026 IEEE. Published in the 2026 IEEE International Geoscience and Remote Sensing Symposium (IGARSS 2026), scheduled for 9 - 14 August 2026 in Washington, D.C.. Personal use of this material is permitted. However, permission to reprint/republish this material for advertising or promotional purposes or for creating new collective works for resale or redistribution to servers or lists, or to reuse any copyrighted component of this work in other works, must be obtained from the IEEE. Contact: Manager, Copyrights and Permissions / IEEE Service Center / 445 Hoes Lane / P.O. Box 1331 / Piscataway, NJ 08855-1331, USA. Telephone: + Intl. 908-562-3966.}
\thanks{This version is the accepted manuscript submitted to arXiv. The final version will be published in the Proceedings of IGARSS 2026 and available via IEEE Xplore. For citation, please refer to the published version in IGARSS 2026.}
}

\author{\IEEEauthorblockN{Zesheng Liu}
\IEEEauthorblockA{\textit{Department of Computer Science and Engineering} \\
\textit{Lehigh University}\\
Bethlehem, USA \\
zel220@lehigh.edu}
\and
\IEEEauthorblockN{Maryam Rahnemoonfar\IEEEauthorrefmark{1}}
\IEEEauthorblockA{\textit{Department of Computer Science and Engineering} \\
\textit{Department of Civil and Environmental Engineering}\\
\textit{Lehigh University}\\
Bethlehem, USA \\
maryam@lehigh.edu}
\thanks{\IEEEauthorrefmark{1} Correspondence to maryam@lehigh.edu}
}

\maketitle
\begin{abstract}
Internal ice layers imaged by radar provide key evidence of snow accumulation and ice dynamics, but radar-derived layer boundary observations are often incomplete, with discontinuous traces and sometimes entirely missing layers, due to limited resolution, sensor noise, and signal loss. Existing graph-based models for ice stratigraphy generally assume sufficiently complete layer profiles and focus on predicting deeper-layer thickness from reliably traced shallow layers. In this work, we address the layer-completion problem itself by synthesizing complete ice-layer thickness annotations from incomplete radar-derived layer traces by conditioning on colocated physical features synchronized from physical climate models. The proposed network combines geometric learning to aggregate within-layer spatial context with a transformer-based temporal module that propagates information across layers to encourage coherent stratigraphy and consistent thickness evolution. To learn from incomplete supervision, we optimize a mask-aware robust regression objective that evaluates errors only at observed thickness values and normalizes by the number of valid entries, enabling stable training under varying sparsity without imputation and steering completions toward physically plausible values. The model preserves observed thickness where available and infers only missing regions, recovering fragmented segments and even fully absent layers while remaining consistent with measured traces. As an additional benefit, the synthesized thickness stacks provide effective pretraining supervision for a downstream deep-layer predictor, improving fine-tuned accuracy over training from scratch on the same fully traced data.
\end{abstract}


\input{V4/introduction.tex}
\input{V4/dataset.tex}

\input{V4/method.tex}
\input{V4/experiment.tex}
\input{V4/results.tex}

\input{V4/conclusion.tex}

% \small
\bibliographystyle{IEEEtran}
\bibliography{references}

\end{document}

%% file: V4/introduction.tex
% !TEX root = ../mainV3.tex

\section{Introduction}

Understanding spatio-temporal patterns in polar ice stratigraphy is important for monitoring snow accumulation, interpreting ice-sheet dynamics, and reducing uncertainty in climate projections. Airborne snow radar systems~\cite{AirborneRadar,snowradar} provide a non-destructive way to observe this structure over large spatial extents by recording radar echograms, where subsurface reflectors appear as layered patterns. However, reliably identifying internal layer boundary traces remains difficult because echograms are noisy, annotations are limited, and deeper reflections are increasingly degraded by attenuation, clutter, low inter-layer contrast, and complex subsurface structure. As a result, layer tracing is often incomplete. Shallow layers are more reliably tracked, whereas deeper layers are faint, fragmented, and sometimes entirely missing. This creates a gap between the scientific value of deep internal layers and the practical difficulty of obtaining dense, reliable labels.

To better model radar-derived ice stratigraphy, Zalatan et al.~\cite{Zalatan_igarss,Zalatan_icip,Zalatan2023} introduced graph-based formulations that represent each internal layer as a spatial graph and predict deep-layer thickness from reliably traced shallow layers. Later works extended this direction with spatio-temporal graph neural networks~\cite{Liu2024_MultiBranch}, graph transformers~\cite{Liu2025_GRIT,Liu2025_STGRIT}, and physics-informed learning strategies~\cite{Liu2024_PIML,Rahnemoonfar2024_IGARSS,Liu2025_RADAR}. However, these methods target downstream thickness prediction under the assumption that sufficiently complete layer profiles are available. They do not directly address the problem considered here: recovering complete layer-thickness annotations when the radar-derived traces themselves are incomplete. In practice, most existing graph-based pipelines therefore require filtering out radargrams that lack enough fully observed layers, which limits their applicability to fragmented real-world measurements.

In this work, we instead address internal ice-layer completion under incomplete layer traces. We propose a physics-conditioned graph transformer that combines geometric information from available traces with colocated physical features synchronized from the Model Atmospheric Regional (MAR) climate model~\cite{MAR_2020,MAR2021}. Each radargram is represented as a sequence of spatial graphs with one graph per internal layer, and the model predicts a thickness value for every node in every layer. Each node includes latitude, longitude, and five MAR-derived physical features describing local accumulation and melt regimes. Because these physical features remain available even when layer traces are fragmented or missing, they provide complementary context for inferring plausible layer geometry where radar returns are weak and annotations are incomplete. To learn from incomplete supervision, we introduce a mask-aware Huber loss that evaluates errors only at observed thickness values and normalizes by the number of valid entries, enabling stable training under varying sparsity without imputation. Unlike prior graph-based ice-stratigraphy models, which assume sufficiently complete inputs for downstream prediction, our model is designed to complete the traces themselves: it preserves observed thickness where available and synthesizes values only at unobserved locations, recovering missing segments and even entirely absent layers while remaining consistent with measured traces.

As a secondary benefit, the synthesized thickness stacks can be used to supervise representation learning in downstream tasks. We pretrain a deep layer-thickness predictor on synthesized stacks and then fine-tune it on fully traced internal layers, achieving higher accuracy and more stable optimization than training the same predictor from scratch.

\begin{figure*}[!h]
  \centering
  \includegraphics[width=1\textwidth]{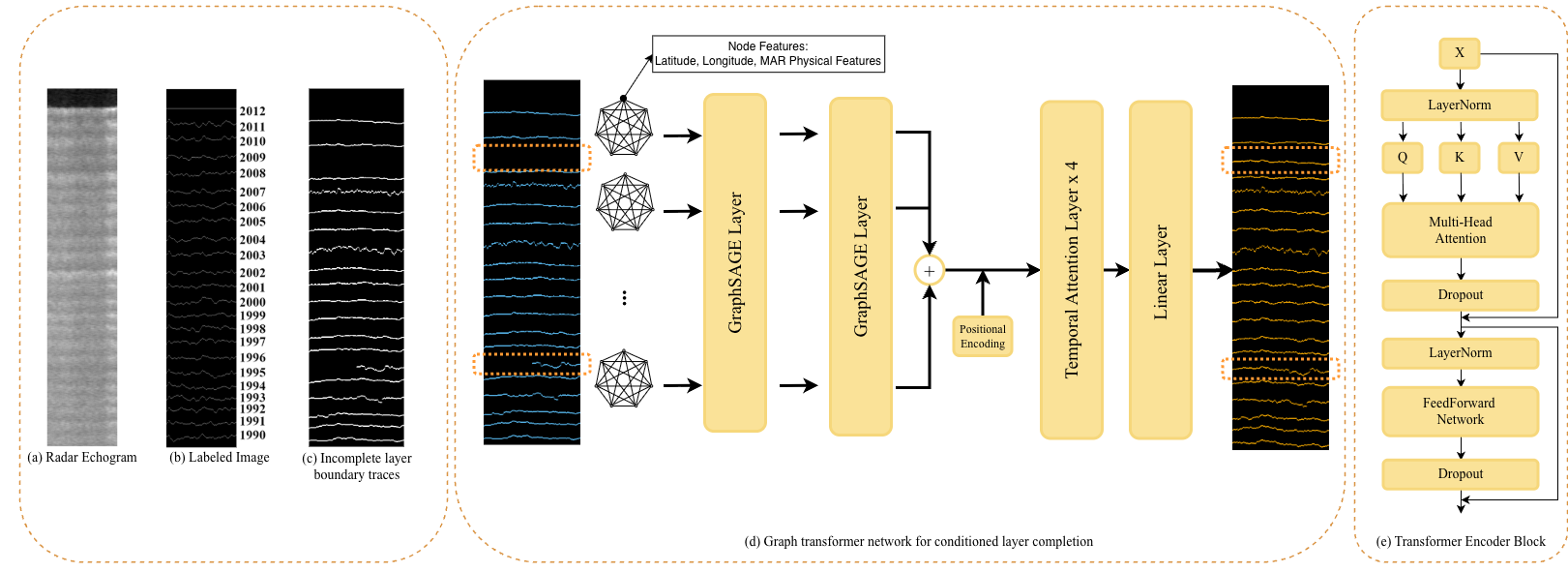}
  \caption{Overview of the dataset and the proposed physics-conditioned graph transformer for layer completion. (\textbf{a}) Radar echogram image. (\textbf{b}) Labeled echogram with manually traced internal layer boundaries. These boundaries can be discontinuous, and some internal layers can be entirely missing. (\textbf{c}) Incomplete internal layer boundary traces, including partial traces and absent layers. (\textbf{d}) Proposed graph network for layer completion from incomplete traces and colocated physical features. (\textbf{e}) Transformer encoder block used in the model.\label{fig:diagram}}
\end{figure*}

%% file: V4/dataset.tex
% !TEX root = ../mainV2.tex

\section{Snow Radar Echogram Dataset}

% \subsection{Snow Radar Echogram Dataset}
In this work, we use the Snow Radar Echogram Dataset (SRED) \cite{ibikunle2025aireadysnowradarechogram}, a large public collection of airborne snow radar echograms for ice layer analysis. The SRED dataset contains approximately 13,800 radargrams acquired over Greenland in 2012 by the CReSIS airborne snow radar as part of NASA’s Operation IceBridge \cite{CReSIS_radar,Leuschen2011SnowRadar}. Airborne snow radar is widely used for cryospheric surveys and ice sheet monitoring because its signals can penetrate thick ice sheets and reflect the internal structure of ice layers in radar echogram images. 

Radar echogram images, as the one shown in Figure \ref{fig:diagram}($\textbf{a}$), encode the strength of the reflected signal at different depths, where brighter pixels indicate stronger reflections \cite{Arnold_2020}. For supervised learning, SRED provides annotated layer boundaries. (Figure \ref{fig:diagram}(\textbf{b})). These boundary traces can be converted into layer thickness profiles by measuring the vertical separation between paired upper and lower layer boundaries at each along-track location. It is worth noting that these layer boundary traces are not always continuous: some layers, as illustrated in Figure~\ref{fig:diagram}(\textbf{c}), can be partially absent or fully missing due to attenuation, noise, or complicated subsurface structure. Such unobserved layer points are recorded as NaN values in the boundary annotations, explicitly marking missing segments and entirely missing layers. The dataset further includes latitude/longitude recorded during acquisition and aligned to the echograms, enabling spatially referenced modeling and fusion with external geophysical covariates.

%% file: V4/method.tex
% !TEX root = ../mainV3.tex

\section{Methodology}

Figure \ref{fig:diagram}(\textbf{d}) illustrates the overall architecture of our proposed graph transformer network for ice layer completion, which combines GraphSAGE inductive geometric learning framework with temporal Transformer encoders to capture both spatial dependencies within each ice layer and temporal dependencies across different layers.

\subsection{Physical Features Provided by the SRED Dataset}
Beyond radar intensity and layer boundary annotations, the SRED dataset~\cite{ibikunle2025aireadysnowradarechogram} provides synchronized physical covariates derived from the Model Atmospheric Regional (MAR) climate model~\cite{MAR_2020, MAR2021}. For each year, SRED aligns MAR v3.10 outputs to the geolocated radar measurements using the recorded latitude/longitude and interpolates gridded MAR fields to radar locations via 2D Delaunay triangulation. As a result, each internal layer is associated with a consistent set of physical features, even when the layer boundary annotations are partially missing or entirely NaN for that year. This is important for our setting, as it provides physically meaningful conditioning information at all nodes despite incomplete radar-derived supervision.

In our method, we use five MAR-derived variables as physical node features: snow mass balance, near-surface temperature, meltwater refreezing, height change due to melting, and snowpack height. These variables are closely linked to surface mass-balance variability and melt/accumulation processes over Greenland~\cite{fettweis2013estimating, vernon2013surface, reeh1991parameterization}, providing useful physical context for thickness synthesis when radar returns are weak and layer boundary traces are discontinuous or entirely missing. To incorporate them into the graph model, we concatenate these physical features with latitude and longitude to form the initial node feature vector for each node.

\subsection{GraphSAGE Inductive Framework}
We adopt GraphSAGE~\cite{Hamilton2018Inductive} as the spatial encoder to model the relational structure within each spatial graph. In contrast to Graph Convolutional Network (GCN)~\cite{kipf2017_GCN}, GraphSAGE is an inductive representation learning approach: it learns neighborhood aggregation functions that can be applied to nodes, and even graphs, not observed during training by combining information from a node's local neighborhood~\cite{ZHOU202057_Review}. In our setting, this means that the encoder operates directly on the input node attributes defined above, including both spatial coordinates and MAR-derived physical variables. A mean-aggregation GraphSAGE layer can be written as
$
\mathbf{x}'_i = \mathbf{W}_1 \mathbf{x}_i + \mathbf{W}_2 \cdot \mathrm{mean}_{j \in \mathcal{N}(i)} \mathbf{x}_j,
$
where $i$ denotes a node, $\mathbf{x}_i$ is its input feature, $\mathbf{x}'_i$ is the updated embedding, $\mathbf{W}_1$ and $\mathbf{W}_2$ are learnable parameters, $\mathcal{N}(i)$ is the neighborhood of node $i$, $\mathbf{x}_j$ denotes features of neighboring nodes, and $\mathrm{mean}(\cdot)$ is the aggregation operator. Because the MAR variables are included in $\mathbf{x}_i$, they participate in every message-passing step. As a result, the learned embedding of each node is conditioned not only on its own local physical state and spatial position, but also on the surrounding physically informed context propagated from neighboring nodes. This design is particularly useful under incomplete layer annotations, since even where radar-derived thickness supervision is missing, the network can still construct meaningful node embeddings from the available coordinates and MAR covariates.

\subsection{Temporal Transformer Encoder}
After the GraphSAGE spatial encoder aggregates spatial information of each layer, we obtain spatial node embeddings $H \in \mathbb{R}^{N \times T \times d_s}$, where $N$ is the number of nodes per layer graph and $T$ is the number of layers in the stack (ordered from shallow to deep, treated as the temporal dimension). We then project $H$ to the Transformer width to obtain $Z \in \mathbb{R}^{N \times T \times d_t}$, and apply a Transformer-style multi-head self-attention encoder \cite{Vaswani2017Attention} along the temporal dimension to propagate information across layers. Figure~\ref{fig:diagram}(\textbf{e}) shows the overall architecture. Given $Z$, we perform the standard multi-head self-attention calculation~\cite{Vaswani2017Attention}. Each encoder layer follows a pre-normalization design, applying LayerNorm before the multi-head self-attention and position-wise feedforward sub-layers, with residual connections and dropout. We use 8 attention heads and stack 4 temporal transformer layers together. Because layer index carries underlying depth information, we use the standard sinusoidal positional encoding described by Vaswani et al. \cite{Vaswani2017Attention} and add it to $Z$ with a layer normalization before feeding into the attention module.

\subsection{Mask-Aware Huber Loss}

To train on incomplete thickness annotations, we use an observation mask and compute the loss only on labeled entries. Let $\mathbf{y}\in\mathbb{R}^{N\times T}$ denote the radar observed ground-truth thickness over $N$ spatial nodes and $T$ layers, and let $\hat{\mathbf{Y}}\in\mathbb{R}^{N\times T}$ be our model prediction. We define a binary mask $\mathbf{M}\in\{0,1\}^{N\times T}$ where $M_{n,t}=1$ if $Y_{n,t}$ is observed and $M_{n,t}=0$ otherwise. With residual $r_{n,t}=\hat{Y}_{n,t}-Y_{n,t}$, our mask-aware Huber loss is the normalized masked average
\begin{equation}
\mathcal{L}_{\text{mask}}=
\frac{1}{\sum_{n=1}^{N}\sum_{t=1}^{T} M_{n,t} + \epsilon}
\sum_{n=1}^{N}\sum_{t=1}^{T} M_{n,t}\,\mathrm{Huber}_\delta(r_{n,t}),
\label{eq:mask_huber}
\end{equation}
where $\mathrm{Huber}_\delta(\cdot)$ is the standard Huber penalty with threshold $\delta$, and $\epsilon$ is a small constant for numerical stability. This formulation ensures that missing entries ($M_{n,t}=0$) contribute neither loss nor gradients, enabling training directly from partially traced layer stacks from radar echograms.

\begin{figure}[t]
  \centering
  \includegraphics[width=0.5\textwidth]{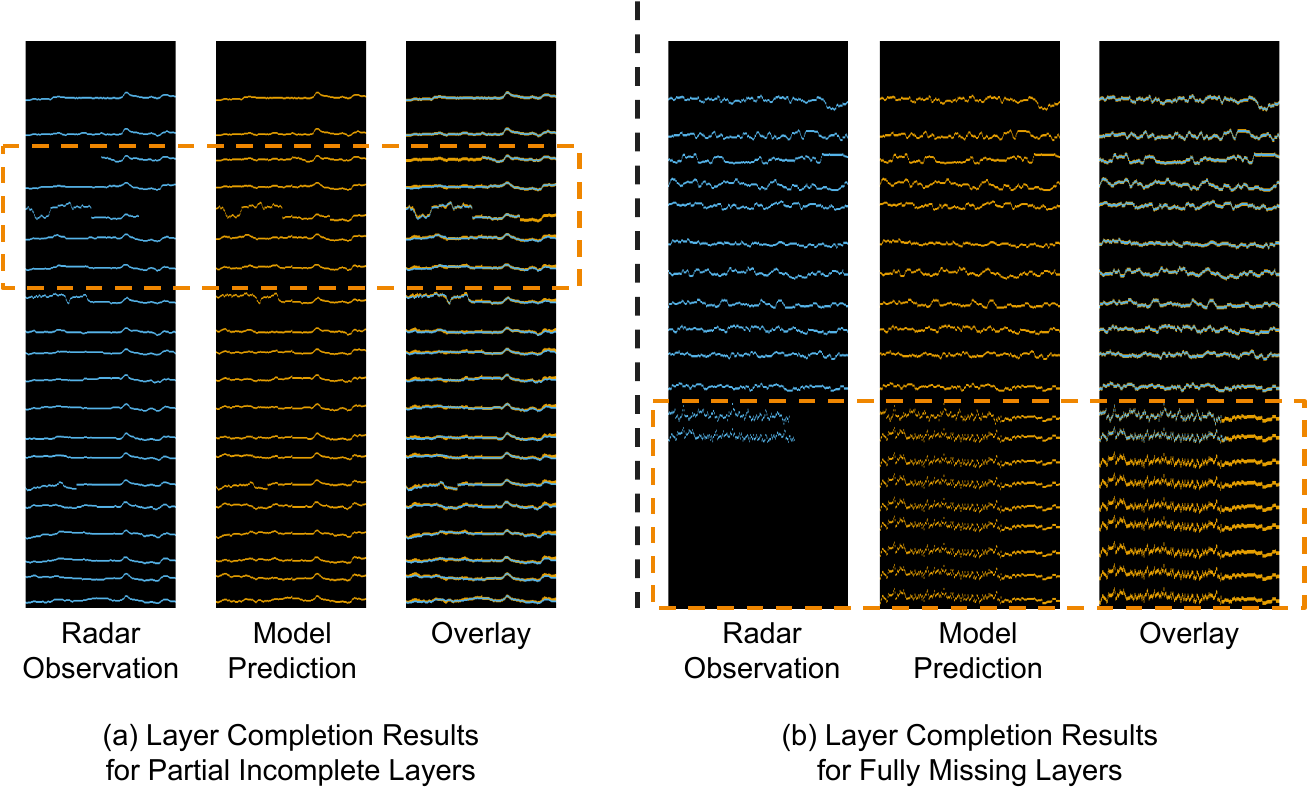}
  \caption{Qualitative examples of our graph-conditioned layer completion model on radar echograms under two annotation-missing regimes. (\textbf{a}) Partially missing traces, where our model fills discontinuities and restores smooth, continuous layer boundaries that remain consistent with the observed segments. (\textbf{b})Fully missing layers, where even when entire internal layers are absent from the input, the model infers coherent layer geometries that follow the overall stratigraphic pattern.  \label{fig:qualitative}} 
\end{figure}

%% file: V4/experiment.tex
% !TEX root = ../mainV3.tex

\section{Experiment Details}
We evalute our proposed graph transformer network on a special case where $T=20$, i.e. we want to complete a stack of 20 internal ice layers for each radar echogram. In the SRED dataset, 1644 radargrams have at least 20 fully tracked internal layers, which we will later use for downstream evaluations. For the remaining 12161 radargrams, we train our proposed layer completion model to complete missing layers up to the 20-th layer. To construct the order sequence of spatial graphs as network input, we follow the same procedure as Zalatan et al.~\cite{Zalatan_igarss,Zalatan_icip,Zalatan2023} and Liu et al.~\cite{Liu2024_MultiBranch}, where each internal layer is represented as a spatial graph with nodes corresponding to pixels in the width of the radar echograms. Each node contains seven node features: Latitude, Longitude, and five physical features from the MAR climate model~\cite{MAR_2020, MAR2021}. 

We implement our graph transformer network in PyTorch and PyTorch Geometric. The spatial encoder is a 2-layer GraphSAGE network with hidden dimension $d_s=128$, and the temporal module is a Transformer encoder with width $d_t=256$, 8 attention heads, and 4 layers. We applied dropout ($p=0.05$) in both GraphSAGE and Transformer blocks. To isolate the contribution of physical conditioning, we also compare our model against a non-physical variant that uses only latitude and longitude as node features. The completion model is trained for 300 epochs with Adam, base learning rate $5\times 10^{-4}$, weight decay $10^{-4}$, and batch size 8, using a mask-aware Huber loss that is computed only on observed thickness values. We use a two-stage learning rate schedule: linear warm-up from 0.1 of the base rate to the base rate over the first 25 epochs, follwed by cosine annealing for the remaining 275 epochs to a minimum learning rate of $10^{-6}$. The final configuration was chosen through preliminary validation experiments informed by prior work and constrained by training stability and GPU memory. 

For downstream evaluation, we adopt the spatio-temporal graph neural network from Liu et al.~\cite{Liu2024_MultiBranch} as the deep layer thickness predictor and follow their training settings for training from scratch on the 1644 fully traced radargrams. For the pretraining-finetuning workflow, we pretrain the same model on synthetic completions from our layer completion model using Liu et al.'s settings for 450 epoch, then fine-tune on the 1644 fully traced radargrams starting from the 100-epoch checkpoint for another 450 epochs with learning rate $7\times 10^{-4}$. During fine-tuning, we use the standard mean squared error (MSE) loss since all layers are fully traced.

%% file: V4/results.tex
% !TEX root = ../mainV3.tex
\section{Results}

\begin{table}[t]
\centering
\caption{Quantitative comparison between the non-physical and physical-conditioned graph transformer variants on the internal ice-layer completion task.\label{table:completion}}
\scalebox{0.9}{
\begin{tabular}{|c|c|c|}
\hline
Model                                  & Masked MAE  \\ \hline
Non-Physical Graph Transformer         &   3.45      \\ \hline
Physical-Conditioned Graph Transfoemer &   \textbf{1.81}   \\ \hline
\end{tabular}
}
\end{table}

Table~\ref{table:completion} compares the graph transformer with and without MAR-derived physical conditioning and calculate the mean absolute error at locations where ground-truth boundary annotations are available. The non-physical variant uses only latitude and longitude as node features, whereas the physical-conditioned variant additionally incorporates MAR-derived physical features. This result highlights the effect of incorporating physical features beyond latitude and longitude for better internal ice-layer completion accuracy. Figure~\ref{fig:qualitative} demonstrates that our method can reliably recover internal layer boundaries from radar echograms even when annotations are severely incomplete. As shown in Figure~\ref{fig:qualitative}(\textbf{a}), when layer traces are only partially available, the predicted boundaries smoothly bridge missing segments while remaining well-aligned with the visible portions of the reference traces in the overlay. Importantly, the completed curves preserve realistic layer ordering and spacing and avoid introducing spurious oscillations, indicating that the model is not performing naïve interpolation but instead leveraging spatial graph connectivity and context from neighboring layers to maintain coherent stratigraphy. In Figure~\ref{fig:qualitative}(\textbf{b}), for the case where entire internal layers are absent from the input, despite having no direct trace evidence for those layers, our model reconstructs a consistent set of boundaries that follow the global geometry of the layer stack and exhibit smooth, physically plausible variations across depth. Taken together, these qualitative examples highlight the effectiveness and robustness of our approach for handling both partial discontinuities and fully missing internal layers, supporting its use as a practical tool for completing sparse layer-boundary annotations.

We further assess whether the synthetic layer traces produced by our method improve downstream learning. To this end, we compare a spatio-temporal GNN~\cite{Liu2024_MultiBranch} pretrained on the synthetic traces and then fine-tuned on radargrams with fully traced internal layers against the same model trained from scratch on the fully traced data. Pretraining leads to faster and more stable optimization during fine-tuning and yields better downstream performance. In particular, the pretrained model achieves an RMSE of 2.9910 pixels for deep-layer thickness prediction, compared with 3.2297 pixels for training from scratch, corresponding to a 7.4\% improvement.

Overall, our physics-conditioned graph transformer effectively synthesizes complete internal ice-layer thicknesses from incomplete layer-boundary annotations, handling both partial discontinuities and fully missing layers while maintaining physically plausible stratigraphy. Furthermore, leveraging these synthetic completions for pretraining enhances downstream deep-layer thickness prediction accuracy and optimization stability, highlighting the practical value of our approach for scalable supervision from partially labeled radargrams.

%% file: V4/conclusion.tex
% !TEX root = ../mainV3.tex

\section{Conclusion}

In this work, we present a novel graph transformer architecture for synthetically completing internal ice layer thicknesses when internal layer boundary annotations are incomplete or discontinuous in radar echograms. Conditioning on climate-model physical features, our model leverages intra-layer spatial structure and cross-layer temporal dependencies to produce physically plausible completions while optimizing a mask-aware Huber loss only on observed measurements. Qualitative results show smooth, coherent, and physically plausible thickness completions that remain consistent with available observations, even under severe missingness. We further demonstrate that these synthetic completions provide effective pretraining supervision: pretraining a downstream spatio-temporal GNN on completed stacks and fine-tuning on fully traced radargrams yields faster convergence, more stable optimization, and a 7.4\% RMSE improvement for deep-layer thickness prediction compared to training from scratch. Overall, this completion-to-pretraining workflow turns partially labeled radargrams into scalable supervision and reduces reliance on fully traced layer stacks for robust deep-layer modeling.

%% file: references.bib
@INPROCEEDINGS{Zalatan_igarss,
  author={Zalatan, Benjamin and Rahnemoonfar, Maryam},
  booktitle={IGARSS 2023 - 2023 IEEE International Geoscience and Remote Sensing Symposium}, 
  title={Prediction of Annual Snow Accumulation Using a Recurrent Graph Convolutional Approach}, 
  year={2023},
  volume={},
  number={},
  pages={5344-5347},
  doi={10.1109/IGARSS52108.2023.10283236}}

@INPROCEEDINGS{Zalatan_icip,
  author={Zalatan, Benjamin and Rahnemoonfar, Maryam},
  booktitle={2023 IEEE International Conference on Image Processing (ICIP)}, 
  title={Prediction of Deep Ice Layer Thickness Using Adaptive Recurrent Graph Neural Networks}, 
  year={2023},
  volume={},
  number={},
  pages={2835-2839},
  doi={10.1109/ICIP49359.2023.10222391}}

@INPROCEEDINGS{Zalatan2023,
  author={Zalatan, Benjamin and Rahnemoonfar, Maryam},
  booktitle={2023 IEEE Radar Conference (RadarConf23)}, 
  title={Recurrent Graph Convolutional Networks for Spatiotemporal Prediction of Snow Accumulation Using Airborne Radar}, 
  year={2023},
  volume={},
  number={},
  pages={1-6},
  doi={10.1109/RadarConf2351548.2023.10149562}}

@misc{Liu2024_PIML,
      title={Learning Spatio-Temporal Patterns of Polar Ice Layers With Physics-Informed Graph Neural Network}, 
      author={Zesheng Liu and Maryam Rahnemoonfar},
      year={2024},
      eprint={2406.15299},
      archivePrefix={arXiv},
      primaryClass={cs.LG},
      url={https://arxiv.org/abs/2406.15299}, 
}

@INPROCEEDINGS{CReSIS_radar,
  author={Gogineni, S. and Yan, J. B. and Gomez, D. and Rodriguez-Morales, F. and Paden, J. and Leuschen, C.},
  booktitle={IEEE MTT-S International Microwave and RF Conference}, 
  title={Ultra-wideband radars for remote sensing of snow and ice}, 
  year={2013},
  volume={},
  number={},
  pages={1-4},
  doi={10.1109/IMaRC.2013.6777743}}

@misc{Leuschen2011SnowRadar,
  author = {Leuschen, Carl and Panzer, Ben and Gogineni, Prasad and Rodriguez, Fernando and Paden, John and Li, Jilu},
  title = {IceBridge Snow Radar L1B Geolocated Radar Echo Strength Profiles},
  year = {2011/2024},  
  howpublished = {Boulder, Colorado USA: National Snow and Ice Data Center. Digital media.},
  note = {Accessed on 2024}  
}

@article{Arnold_2020, title={CReSIS airborne radars and platforms for ice and snow sounding}, volume={61}, DOI={10.1017/aog.2019.37}, number={81}, journal={Annals of Glaciology}, author={Arnold, Emily and Leuschen, Carl and Rodriguez-Morales, Fernando and Li, Jilu and Paden, John and Hale, Richard and Keshmiri, Shawn}, year={2020}, pages={58–67}}

@misc{Hamilton2018Inductive,
      title={Inductive Representation Learning on Large Graphs}, 
      author={William L. Hamilton and Rex Ying and Jure Leskovec},
      year={2018},
      eprint={1706.02216},
      archivePrefix={arXiv},
      primaryClass={cs.SI}
}

@article{ZHOU202057_Review,
title = {Graph neural networks: A review of methods and applications},
journal = {AI Open},
volume = {1},
pages = {57-81},
year = {2020},
issn = {2666-6510},
doi = {https://doi.org/10.1016/j.aiopen.2021.01.001},
url = {https://www.sciencedirect.com/science/article/pii/S2666651021000012},
author = {Jie Zhou and Ganqu Cui and Shengding Hu and Zhengyan Zhang and Cheng Yang and Zhiyuan Liu and Lifeng Wang and Changcheng Li and Maosong Sun}
}

@misc{kipf2017_GCN,
      title={Semi-Supervised Classification with Graph Convolutional Networks}, 
      author={Thomas N. Kipf and Max Welling},
      year={2017},
      eprint={1609.02907},
      archivePrefix={arXiv},
      primaryClass={cs.LG},
      url={https://arxiv.org/abs/1609.02907}, 
}

@incollection{AirborneRadar,
title = {Chapter 12 - Glaciers and Ice Sheets},
editor = {Harry M. Jol},
booktitle = {Ground Penetrating Radar Theory and Applications},
publisher = {Elsevier},
address = {Amsterdam},
pages = {361-392},
year = {2009},
isbn = {978-0-444-53348-7},
doi = {https://doi.org/10.1016/B978-0-444-53348-7.00012-0},
url = {https://www.sciencedirect.com/science/article/pii/B9780444533487000120},
author = {Steven A. Arcone}
}

@INPROCEEDINGS{snowradar,
  author={Gogineni, S. and Yan, J. B. and Gomez, D. and Rodriguez-Morales, F. and Paden, J. and Leuschen, C.},
  booktitle={IEEE MTT-S International Microwave and RF Conference}, 
  title={Ultra-wideband radars for remote sensing of snow and ice}, 
  year={2013},
  volume={},
  number={},
  pages={1-4},
  doi={10.1109/IMaRC.2013.6777743}}

@article{vaswani2017attention,
  title={Attention is all you need},
  author={Vaswani, A},
  journal={Advances in Neural Information Processing Systems},
  year={2017}
}

@INPROCEEDINGS{Rahnemoonfar2024_IGARSS,
  author={Rahnemoonfar, Maryam and Zalatan, Benjamin},
  booktitle={IGARSS 2024 - 2024 IEEE International Geoscience and Remote Sensing Symposium}, 
  title={Physics-informed Machine Learning for Deep Ice Layer Tracing in SAR images}, 
  year={2024},
  volume={},
  number={},
  pages={6938-6942},
  doi={10.1109/IGARSS53475.2024.10641831}}

@misc{ibikunle2025aireadysnowradarechogram,
      title={AI-ready Snow Radar Echogram Dataset (SRED) for climate change monitoring}, 
      author={Oluwanisola Ibikunle and Hara Talasila and Debvrat Varshney and Jilu Li and John Paden and Maryam Rahnemoonfar},
      year={2025},
      eprint={2505.00786},
      archivePrefix={arXiv},
      primaryClass={cs.CV},
      url={https://arxiv.org/abs/2505.00786}, 
}

@article{fettweis2013estimating,
  title={Estimating the Greenland ice sheet surface mass balance contribution to future sea level rise using the regional atmospheric climate model MAR},
  author={Fettweis, Xavier and Franco, Bruno and Tedesco, Marco and Van Angelen, JH and Lenaerts, Jan TM and van den Broeke, Michiel R and Gall{\'e}e, H},
  journal={The Cryosphere},
  volume={7},
  number={2},
  pages={469--489},
  year={2013},
  publisher={Copernicus GmbH}
}

@article{vernon2013surface,
  title={Surface mass balance model intercomparison for the Greenland ice sheet},
  author={Vernon, Christopher L and Bamber, JL and Box, JE and Van den Broeke, MR and Fettweis, Xavier and Hanna, Edward and Huybrechts, Phillipe},
  journal={The Cryosphere},
  volume={7},
  number={2},
  pages={599--614},
  year={2013},
  publisher={Copernicus Publications G{\"o}ttingen, Germany}
}

@article{reeh1991parameterization,
  title={Parameterization of melt rate and surface temperature in the Greenland ice sheet},
  author={Reeh, Niels},
  journal={Polarforschung},
  volume={59},
  number={3},
  pages={113--128},
  year={1991},
  publisher={Alfred Wegener Institute for Polar and Marine Research \& German Society of~…}
}

@Article{MAR_2020,
AUTHOR = {Delhasse, A. and Kittel, C. and Amory, C. and Hofer, S. and van As, D. and S. Fausto, R. and Fettweis, X.},
TITLE = {Brief communication: Evaluation of the near-surface climate in ERA5 over the Greenland Ice Sheet},
JOURNAL = {The Cryosphere},
VOLUME = {14},
YEAR = {2020},
NUMBER = {3},
PAGES = {957--965},
URL = {https://tc.copernicus.org/articles/14/957/2020/},
DOI = {10.5194/tc-14-957-2020}
}

@Article{MAR2021,
AUTHOR = {Fettweis, X. and Hofer, S. and S\'ef\'erian, R. and Amory, C. and Delhasse, A. and Doutreloup, S. and Kittel, C. and Lang, C. and Van Bever, J. and Veillon, F. and Irvine, P.},
TITLE = {Brief communication: Reduction in the future
Greenland ice sheet surface melt with the help of solar geoengineering},
JOURNAL = {The Cryosphere},
VOLUME = {15},
YEAR = {2021},
NUMBER = {6},
PAGES = {3013--3019},
URL = {https://tc.copernicus.org/articles/15/3013/2021/},
DOI = {10.5194/tc-15-3013-2021}
}

@misc{Liu2024_MultiBranch,
      title={Multi-branch Spatio-Temporal Graph Neural Network For Efficient Ice Layer Thickness Prediction}, 
      author={Zesheng Liu and Maryam Rahnemoonfar},
      year={2024},
      eprint={2411.04055},
      archivePrefix={arXiv},
      primaryClass={cs.LG},
      url={https://arxiv.org/abs/2411.04055}, 
}

@INPROCEEDINGS{Liu2025_STGRIT,
  author={Liu, Zesheng and Rahnemoonfar, Maryam},
  booktitle={2025 IEEE International Conference on Image Processing (ICIP)}, 
  title={ST-GRIT: Spatio-Temporal Graph Transformer For Internal Ice Layer Thickness Prediction}, 
  year={2025},
  volume={},
  number={},
  pages={1109-1114},
  doi={10.1109/ICIP55913.2025.11084445}}

@INPROCEEDINGS{Liu2025_RADAR,
  author={Liu, Zesheng and Rahnemoonfar, Maryam},
  booktitle={2025 IEEE International Radar Conference (RADAR)}, 
  title={Physics-Informed Spatio-Temporal Graph Neural Network for Efficient Deep Ice Layer Thickness Estimation in Radar Imagery}, 
  year={2025},
  volume={},
  number={},
  pages={1-6},
  doi={10.1109/RADAR52380.2025.11031955}}

@INPROCEEDINGS{Liu2025_GRIT,
  author={Liu, Zesheng and Rahnemoonfar, Maryam},
  booktitle={IGARSS 2025 - 2025 IEEE International Geoscience and Remote Sensing Symposium}, 
  title={GRIT: Graph Transformer For Internal Ice Layer Thickness Prediction}, 
  year={2025},
  volume={},
  number={},
  pages={1-5},
  doi={10.1109/IGARSS55030.2025.11243115}}
